\pdfoutput=1

\documentclass[11pt]{article}

\usepackage{acl}

\usepackage{times}
\usepackage{latexsym}

\usepackage[T1]{fontenc}

\usepackage[utf8]{inputenc}

\usepackage{microtype}

\usepackage{times}
\usepackage{latexsym}
\usepackage{subcaption}
\usepackage{xcolor}
\usepackage{graphicx}
\usepackage{array}
\usepackage{amsmath}
\usepackage{amssymb}
\usepackage{bbm}
\usepackage{booktabs}
\usepackage{longtable}
\usepackage{colortab}
\usepackage{colortbl}
\usepackage{arydshln}
\usepackage{tabularx}
\usepackage{setspace}
\usepackage{mathtools}
\usepackage{multirow}
\usepackage{eqnarray}
\usepackage{seqsplit}
\usepackage{makecell}

\usepackage{url}
\usepackage{ascmac}

\usepackage[hang,flushmargin]{footmisc} 

\title{KART: Parameterization of Privacy Leakage Scenarios from Pre-trained Language Models}

\author{
    Yuta Nakamura$^{1}$,
    Shouhei Hanaoka$^{1,2}$,
    Yukihiro Nomura$^{3,4}$, Naoto Hayashi$^{3}$, 
    \\
    \textbf{
    Osamu Abe$^{1,2}$, Shuntaro Yada$^{5}$, Shoko Wakamiya$^{5}$, Eiji Aramaki$^{5}$
    }
    \\
    
    \text{$^{1}$The University of Tokyo}
    
    \text{$^{2}$The Department of Radiology, The University of Tokyo Hospital}\\
    
    \text{$^{3}$The Department of Computational Diagnostic Radiology and Preventive Medicine, The University of Tokyo Hospital}\\
    
    \text{$^{4}$Chiba University}
    
    \text{$^{5}$Nara Institute of Science and Technology}
    
    \vspace{-0.5ex}
    \\
    \small{
        \texttt{\{yutanakamura-tky,hanaoka-tky,nomuray-tky,naoto-tky,abediag-tky\}@umin.ac.jp}
    }
    \vspace{-0.8ex}
    \\
    \small{
        \texttt{ynomura@chiba-u.jp}
    }
    \vspace{-1ex}
    \\
    \small{
        \texttt{\{s-yada,wakamiya,aramaki\}@is.naist.jp}
    }
}

\date{}

\begin{document}
\maketitle

\begin{abstract}
For the safe sharing pre-trained language models, no guidelines exist at present owing to the difficulty in estimating the upper bound of the risk of privacy leakage. One problem is that previous studies have assessed the risk for different real-world privacy leakage scenarios and attack methods, which reduces the portability of the findings.
To tackle this problem, we represent complex real-world privacy leakage scenarios under a universal parameterization, \textit{Knowledge, Anonymization, Resource, and Target} (KART). KART parameterization has two merits: (i) it clarifies the definition of privacy leakage in each experiment and (ii) it improves the comparability of the findings of risk assessments. We show that previous studies can be simply reviewed by parameterizing the scenarios with KART. We also demonstrate privacy risk assessments in different scenarios under the same attack method, which suggests that KART helps approximate the upper bound of risk under a specific attack or scenario. We believe that KART helps integrate past and future findings on privacy risk and will contribute to a standard for sharing language models.
\end{abstract}

\vspace{-1ex}

\begin{figure*}[tb]
    \begin{center}
        \includegraphics[scale=0.30]{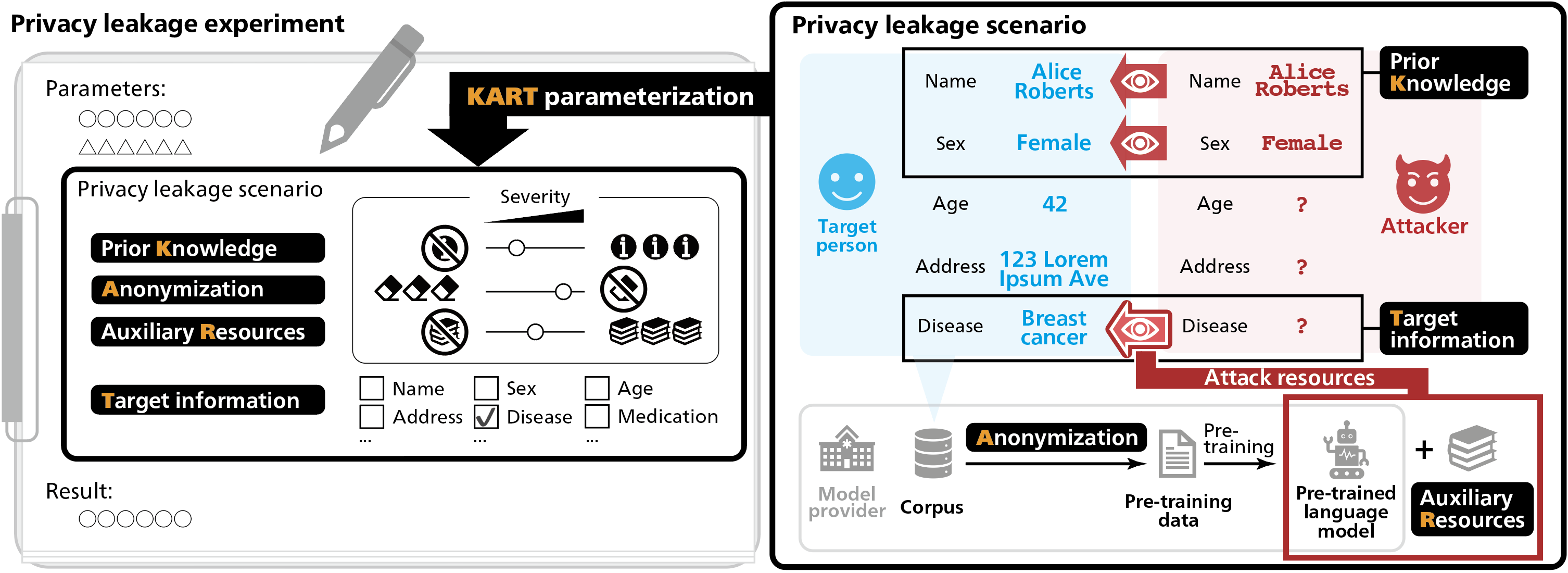}
        \vspace{-4ex}
        \caption{
Scenario-aware privacy risk assessment using KART parameterization. Any privacy leakage experiment implicitly or explicitly assumes the scenario where an attacker who has \textit{prior knowledge} about a target person attacks a pre-trained language model to obtain other personal \textit{target information}. The attacker may also use \textit{auxiliary resources}. The target information may or may not be in the pre-training data depending on the \textit{anonymization}. Parameterizing the assumed scenario would improve the portability of the findings of the experiment.}
        \label{fig:overview_1}
    \end{center}
    \vspace{-3ex}
\end{figure*}

\section{Introduction}
Recent natural language processing (NLP) has benefited from language models such as Transformer~\cite{transformer}, GPT~\cite{gpt}, and BERT~\cite{bert}. However, we face a privacy risk when sharing language models since personal information in the pre-training data could be recovered from the models~\cite{membership_inference_gpt_misra,membership_inference_seq2seq,exposure_and_unintended_memorization,extracting_training_data_from_lm,training-data-leakage-analysis-in-language-models-inan,bert-privacy-naacl-2021-lehman,bert-privacy-vakili,risk-evaluation-of-differential-privacy-emnlp-2021-hoory-shlomo}.

No guidelines for publishing pre-trained language models have been established since we lack knowledge about the impact of a model release on privacy safety. This is in contrast to the data itself, for which standards of processing, sharing, and publishing have already been legislated in several countries~\cite{hipaa,gdpr}. Alternatively, model providers have published language models only when the pre-training data is free of sensitive personal information. In the biomedical domain, for example, BioBERT~\cite{biobert} and BioMegatron~\cite{biomegatron} are publicly available, both of which are pre-trained with biomedical articles. ClinicalBERT~\cite{clinicalbert}, BlueBERT~\cite{bluebert}, UTH-BERT~\cite{uthbert}, and MS-BERT~\cite{ms-bert}, which use manually anonymized clinical records, have also been published. However, EhrBERT~\cite{ehrbert} and AlphaBERT~\cite{alphabert} are also pre-trained with clinical records but have not been released.

How can we decide whether a language model is safe enough to share? Studies have assessed the privacy risk under various attacks and real-world privacy leakage scenarios, but it has been difficult to integrate them to estimate the upper bound of the risk. Moreover, no studies have directly compared the risk under the same definition of privacy leakage with different attack methods or scenarios. We attribute these shortcomings to a lack of methodology to clarify the presupposed scenarios.

To address the issue, we represent a privacy leakage scenario as a set of primary factors under the \textit{Knowledge, Anonymization, Resource, and Target} (KART) parameterization, as in Figure \ref{fig:overview_1}. The primary factors are as follows: (i) \textit{Prior knowledge} (K): information of the target people already known to the attacker;  (ii) \textit{Target information} (T): personal information that the attacker wishes to obtain; (iii) \textit{Anonymization} (A): removal of personal information from the pre-training data; (iv) \textit{Auxiliary resources} (R): resources used by the attacker other than the language model. We show that KART can simplify complex scenarios assumed in previous studies and assist risk comparison in different scenarios or attacks.

Our contribution is an introduction of KART, which will enable comprehensive privacy risk assessment by improving scenario awareness and the portability of findings of past and future studies.

\section{Related Work}
\subsection{Security of clinical records}
Data is not secure after only deleting attributes that alone can determine from whom the data originated, such as names or IDs.
The data can still be re-identified using remaining attributes and external data.~\citet{k-anonymity} proposed the deletion or generalization of attributes to prevent re-identification.

Clinical records must be handled carefully as they contain sensitive health information that patients do not wish to spread unnecessarily~\cite{ehr_privacy_review,bigdata_review}. Improper disclosure may lead to mental pain, biases in education and employment, and target marketing to vulnerable people~\cite{privacy_medical_big_data}. In the United States, the research use of health information is mainly regulated by the Federal Common Rule for human subject research and the Health Insurance Portability and Accountability Act of 1996 (HIPAA). The HIPAA Privacy Rule\footnote{\url{https://www.hhs.gov/hipaa/for-professionals/privacy/index.html}} refers to sensitive clinical information as protected health information (PHI), such as clinical history, clinical test results, and genomes. General identifiers such as names and addresses are also PHI if linked with health information. The HIPAA Privacy Rule obligates the 18 identifiers listed in Appendix \ref{sec:18-identifiers} to be removed from clinical records for the second usage~\cite{hipaa}.

\subsection{Privacy attacks on language models}
There are two types of prior study on the privacy risk in NLP according to the attack method.

The first is on attacks to obtain personal information in input texts from gradients or encoded representations~\cite{deep_leakage_from_gradient,information_leakage_embedding,risk_of_language_model}. The second is on attacks to restore personal information in the pre-training data from a model, which we review further in Section \ref{sec:kart-parameterization-of-previous-studies}. They are divided into membership inference and model inversion.
Membership inference is a prediction of whether a document was in the pre-training data~\cite{membership_inference,membership_inference_gpt_misra,membership_inference_seq2seq}. Model inversion is an estimation of specific attributes of target people~\cite{model_inversion,exposure_and_unintended_memorization,extracting_training_data_from_lm,training-data-leakage-analysis-in-language-models-inan,bert-privacy-naacl-2021-lehman,bert-privacy-vakili,risk-evaluation-of-differential-privacy-emnlp-2021-hoory-shlomo}.

\renewcommand{\arraystretch}{0.85}
\begin{table*}[tb]
    \begin{center}
    \begin{small}
        \centering
        \caption{Examples of the possible values of the \textit{K}, \textit{A}, \textit{R}, and \textit{T} factors in KART parameterization.}
    \vspace{-2.5ex}
    
    \begin{tabular}{cll}\toprule
    Factor & Value & Scenario
    \\ \midrule
    \multirow{2}{*}{\vspace{-1ex}\textit{K}} & $\varnothing$ & The attacker has no prior knowledge
    \\ \cmidrule{2-3}
    & $\{(\textit{age}, \textit{sex})_{p}\}_{p\in\mathcal{P}}$ & The attacker already knows the age and sex of the target people
    
    \\ \midrule
    
    \multirow{2}{*}{\vspace{-1ex}\textit{A}} & $f_{\text{HIPAA}}$ & The pre-training data is anonymized under the HIPAA Privacy Rule
    \\ \cmidrule{2-3}
    & $\mathrm{id}$ & The pre-training data is not anonymized at all
    
    \\ \midrule
    
    \multirow{3}{*}{\vspace{-3ex}\textit{R}} & 
    $\varnothing$
    & 
    \hspace{-2ex}
        \begin{tabular}{l}
         The attacker has no resource other than $\mathcal{M}$
        \end{tabular}

    \\ \cmidrule{2-3}
    &
    $\{
    \widetilde{\mathcal{D}}_{\text{train}} \}$
    & 
    \hspace{-2ex}
        \begin{tabular}{l}
         The attacker has obtained a corpus with a similar distribution to the pre-training data
        \end{tabular}
 
    \\ \cmidrule{2-3}
    &
    $
    \{\textit{full name}_{\overline{p}}\}_{\overline{p}\in\overline{\mathcal{P}}}
    $
    & 
    \hspace{-2ex}
        \begin{tabular}{l}
         The attacker knows the full name of the non-target people in the pre-training data
        \end{tabular}

    \\ \midrule
    
    \multirow{3}{*}{\vspace{-7ex}\textit{T}} & $\{\textit{address}_{p}\}_{p \in \mathcal{P}}$ &
    The attacker predicts the address of the target people (model inversion)
    \\ \cmidrule{2-3}
    & $\{(\textit{full name}, \textit{address})_{p}\}_{p \in \mathcal{P}}$
    & 
    \hspace{-2ex}
    \begin{tabular}{l}
     The attacker predicts the full name and address of the target people together \\ (model inversion)
    \end{tabular}
    \\ \cmidrule{2-3}
    & $\{\mathbbm{1}[d\in\mathcal{D}_{\text{train}}]\}_{d\in \mathcal{D}}$
    & 
        \hspace{-2ex}
        \begin{tabular}{l}
         The attacker predicts whether a document $d \in \mathcal{D}$ was in the pre-training data \\
         (membership inference)
        \end{tabular}
    
    \\ \bottomrule

    \end{tabular}
    \label{tb:values_of_k_a_r_t}
    \end{small}
    \end{center}
    \vspace{-2.5ex}
\end{table*}
\renewcommand{\arraystretch}{1.0}

\subsection{Generalization in privacy risk evaluation}
It is difficult to estimate the upper bound of the privacy risk from pre-trained language models covering numerous privacy leakage scenarios. Several universal evaluation methods have been introduced.

\citet{exposure_and_unintended_memorization} proposed \textit{exposure}, the frequency with which natural language generation (NLG) reproduces canary sequences added to the pre-training data.~\citet{training-data-leakage-analysis-in-language-models-inan} used the perplexity with which a model $\mathcal{M}$ outputs exact substrings of the pre-training data. If the substrings derive from a single person and $\mathcal{M}$ gives a far lower perplexity than other models, the disclosure can be actual privacy leakage rather than coincidental.

Differential privacy~\cite
{differential_privacy} is a constraint on privacy risk integrated with privacy mechanisms. In machine learning, a model $\mathcal{M}$ pre-trained on a dataset $D$ is $\epsilon$-differentially private if, for any adjacent dataset $D'$ different from $D$ in a single record, the probability that a model $\mathcal{M}'$ pre-trained on $D'$ is distinguished from $\mathcal{M}$ never exceeds the upper bound defined by $\epsilon$. Model providers can determine the value of $\epsilon$ beforehand and ensure privacy by perturbing a model correspondingly during training~\cite{deep-learning-with-differential-privacy-abadi-martin,differential-privacy-in-language-models-kerrigan-2020}. Differential privacy is mathematically guaranteed to be robust to any prior knowledge of the attacker.~\citet{risk-evaluation-of-differential-privacy-emnlp-2021-hoory-shlomo} assessed the practical privacy risk of an $\epsilon$-differentially private language model using the \textit{exposure} metric~\cite{exposure_and_unintended_memorization}.

\section{KART Parameterization}
We propose KART to characterize a privacy leakage scenario with four primary factors. Refer to Table \ref{tb:values_of_k_a_r_t} for examples of the values of each factor.

\vspace{1ex}

\noindent \textbf{Personal information} We denote personal information as $\textit{category}_{\textit{person}} = \textit{value}$.
For example, $\text{``}\textit{full name}_{p_{0}} = \text{Alice Roberts''}$ means that the full name of the person $p_{0}$ is Alice Roberts.
We denote the universal set of all personal information in the world as $I_{U}$: $I_{U} =\{\textit{category}_{\textit{person}}\}_{\forall{\textit{person}}, \forall{\textit{category}}}$.

Let $\mathcal{D}_{\text{private}}$ denote a corpus, all of whose documents are used for pre-training, and let $\mathcal{P}$ be the set of people in $\mathcal{D}_{\text{private}}$. We denote the set of the personal information in $\mathcal{D}_{\text{private}}$ as $I_{\mathcal{D}_{\text{private}}}$  $(\subset I_{U})$.

\vspace{1ex}

\noindent \textbf{\textit{A} factor: anonymization} The \textit{A} factor is an operation, which we denote as $a$, to anonymize $\mathcal{D}_{\text{private}}$ to build pre-training data. Examples of $a$ are the complete manual deletion of personal information under the HIPAA Privacy Rule $(a=f_{\text{HIPAA}})$, automated anonymization, and no operation at all as long as the model is strictly kept private $(a=\mathrm{id})$.

We denote the pre-training data as $\mathcal{D}_{\text{train}} = a(\mathcal{D}_{\text{private}})$ and the set of the remaining personal information as $I_{\mathcal{D}_{\text{train}}} = a(I_{\mathcal{D}_{\text{private}}})$.

Let $\mathcal{M}$ be the pre-trained language model. $\mathcal{M}$ may memorize some of the personal information in $a(I_{\mathcal{D}_{\text{private}}})$ during the pre-training. We denote such memorization as $m$ and the set of the memorized information as $m(a(I_{\mathcal{D}_{\text{private}}}))$. $m(a(I_{\mathcal{D}_{\text{private}}}))$ is a subset of $a(I_{\mathcal{D}_{\text{private}}})$, which is a subset of $I_{\mathcal{D}_{\text{private}}}$: $m(a(I_{\mathcal{D}_{\text{private}}})) \subseteq a(I_{\mathcal{D}_{\text{private}}}) \subseteq I_{\mathcal{D}_{\text{private}}}$.

The attacker can obtain all of $m(a(I_{\mathcal{D}_{\text{private}}}))$ in the worst case, but the personal information not memorized by $\mathcal{M}$ will not leak.

\vspace{1ex}

\noindent \textbf{\textit{K} and \textit{T} factors: prior knowledge and target information}
The \textit{K} factor is the set of prior knowledge about the target people that is already known to the attacker such as full name, age, or sex. We denote the set as $I_{K}$. $I_{K}$ is a subset of $I_{U}$: $I_{K} \subseteq I_{U}$.

The \textit{T} factor is the set of personal information that the attacker aims to obtain, which we note as $I_{T}$. This greatly affects the definition of the privacy risk since it determines which pieces of information in the pre-training data are considered in the privacy leakage. $I_{K}$ and $I_{T}$ are disjoint: $I_{K} \cap I_{T} = \varnothing$.

In model inversion, $I_{T}$ is a subset of $I_{U}$: $I_{T} \subseteq I_{U}$. In membership inference, $I_{T}$ is existence or absence of arbitrary documents in the pre-training data: $I_{T}=\{\mathbbm{1}[d\in\mathcal{D}_{\text{train}}]\}_{d\in \mathcal{D}}$.

The notion of the \textit{K} factor rests on the fact that the more prior knowledge available on a person, the more identifiable the person becomes after privacy attacks. Suppose that the attacker aims to obtain the disease names of a person $p_{1}$ and has revealed that ``diabetes'' was in the pre-training data:

\vspace{-0.5ex}

\begin{center}
$
I_{T}\!=\!\{\textit{diseases}_{p_{1}}\},\;
\{\text{diabetes}\}\!\subset\! m(a(I_{\mathcal{D}_{\text{private}}}))
$.
\end{center}

\vspace{-0.4ex}

\noindent The attacker does not yet know the diseases of $p_{1}$ since it is unclear who is diabetic in the pre-training data. However, if the attacker knows the full name of $p_{1}$ $(I_{K}=\{\textit{full name}_{p_{1}}\})$ and the model $\mathcal{M}$ associates ``diabetes'' with the full name, the attacker can infer that $p_{1}$ is diabetic. This is expressed as below with $\widehat{x}$ denoting the prediction of $x$:

\vspace{-0.4ex}

\begin{center}

$
\widehat{\textit{diseases}_{p_{1}}}\!=\! \{\text{diabetes}\},\:\:
\widehat{I_{T}}\!=\! \{\widehat{\textit{diseases}_{p_{1}}}\}
$.

\end{center}

\vspace{-0.4ex}

Even if the attacker has prior knowledge about people who are \textit{not} the target of the attack, we do not include such knowledge in the \textit{K} factor but the \textit{R} factor, as discussed in the next subsection.

\renewcommand{\arraystretch}{0.85}
\begin{table*}[tb]
    \begin{center}
    \begin{small}
        \centering
        \caption{
        KART parameterization of the simulated privacy leakage scenarios in previous studies.
        }
    \vspace{-2.5ex}
    \begin{tabular}{llllll}\toprule
    Study & $I_{K}$ & $a$ & $R$ & $I_{T}$ & Attack method
    
    \\ \midrule
    
    \citet{membership_inference_gpt_misra}
    & $\varnothing$
    & $\mathrm{id}$
    & $\{\mathcal{D}_{\text{train}}^{\overline{\mathcal{P}}}\}$
    & $\{\mathbbm{1}[d\in\mathcal{D}_{\text{train}}]\}_{d\in\mathcal{D}}$
    & Membership inference
    
    \\ \midrule
    
    \citet{membership_inference_seq2seq}
    & $\varnothing$
    & $\mathrm{id}$
    & $\{\mathcal{D}_{\text{train}}^{\overline{\mathcal{P}}}\}$
    & $\{\mathbbm{1}[d\in\mathcal{D}_{\text{train}}]\}_{d\in\mathcal{D}}$
    & Membership inference
    
    \\ \midrule
    
    \multirow{2}{*}{\vspace{-2ex}\citet{exposure_and_unintended_memorization}}
    & $\varnothing$
    & $\mathrm{id}$
    & $\varnothing$
    &
    \hspace{-2ex}
    \begin{tabular}{l}
        $\{\textit{credit card}$\\ $\:\:\textit{number}_{p}\}_{p\in\mathcal{P}}$
    \end{tabular}
    & NLG$^{*}$
    \\
    & $\varnothing$
    & $\mathrm{id}$
    & $\varnothing$
    &
    \hspace{-2ex}
    \begin{tabular}{l}
        $\{\textit{social security}$\\ $\:\:\textit{number}_{p}\}_{p\in\mathcal{P}}$
    \end{tabular}
    & NLG$^{*}$
    
    \\ \midrule
    
    \citet{extracting_training_data_from_lm}
    & $\varnothing$
    & $\mathrm{id}$
    & $\varnothing$
    & $I_{U}$
    & NLG$^{*}$
    
    \\ \midrule
    
    \citet{training-data-leakage-analysis-in-language-models-inan}
    & $\varnothing$
    & $\mathrm{id}$
    & $\varnothing$
    & $I_{U}$
    & NLG$^{*}$

    \\ \midrule

    \multirow{5}{*}{\vspace{-3ex}\citet{bert-privacy-naacl-2021-lehman}}
    &
    \hspace{-2ex}
    \begin{tabular}{l}
        $\{(\textit{full name}, $\\ $\:\:\:\:\textit{sex})_{p}\}_{p\in\mathcal{P}}$
    \end{tabular}
    & $\mathrm{id}$
    & $\varnothing$
    & $\{\textit{diseases}_{p}\}_{p\in\mathcal{P}}$
    & Language modeling$^{*}$
    \\
    &
    \hspace{-2ex}
    \begin{tabular}{l}
        $\{(\textit{full name}, $\\ $\:\:\:\:\textit{sex})_{p}\}_{p\in\mathcal{P}}$
    \end{tabular}
    & $\mathrm{id}$
    &
    \hspace{-2ex}
    \begin{tabular}{l}
        $\{(\textit{full name}, \textit{sex}, $\\ $\:\:\:\:\textit{diseases})_{\overline{p}}\}_{\overline{p}\in\overline{\mathcal{P}}}$
    \end{tabular}
    & $\{\textit{diseases}_{p}\}_{p\in\mathcal{P}}$
    & Classification$^{*}$
    \\
    & $\varnothing$
    & $\mathrm{id}$
    & $
    \{\textit{full name}_{\overline{p}}\}_{\overline{p}\in\overline{\mathcal{P}}}
    $
    & $\{\textit{full name}_{p}\}_{p\in\mathcal{P}}$
    & Classification$^{*}$
    \\
    & $\{\textit{first name}_{p}\}_{p\in\mathcal{P}}$
    & $\mathrm{id}$
    & $\varnothing$
    & $\{\textit{last name}_{p}\}_{p\in\mathcal{P}}$
    & Language modeling$^{*}$
    \\
    & $\{\textit{last name}_{p}\}_{p\in\mathcal{P}}$
    & $\mathrm{id}$
    & $\varnothing$
    & $\{\textit{first name}_{p}\}_{p\in\mathcal{P}}$
    & Language modeling$^{*}$
    
    \\ \midrule
    
    \multirow{2}{*}{\vspace{-2ex}\citet{bert-privacy-vakili}}
    & $\varnothing$
    & $\mathrm{id}$
    & $\varnothing$
    &
    \hspace{-2ex}
    \begin{tabular}{l}
        $\{(\textit{full name}, $\\ $\:\:\:\:\textit{diseases})_{\overline{p}}\}_{\overline{p}\in\overline{\mathcal{P}}}$
    \end{tabular}
    & NLG$^{*}$
    \\
    & $\{\textit{sex}_{p}\}_{p\in\mathcal{P}}$
    & $\mathrm{id}$
    & $\varnothing$
    &
        \hspace{-2ex}
    \begin{tabular}{l}
        $\{(\textit{full name}, $\\ $\:\:\:\:\textit{diseases})_{\overline{p}}\}_{\overline{p}\in\overline{\mathcal{P}}}$
    \end{tabular}
    & NLG$^{*}$
    \\
    \bottomrule
    \end{tabular}
    
    \vspace{-1.1ex}
    
    \begin{flushleft}
    $^{*}$ Model inversion.
    \end{flushleft}
    
    \label{tb:kart-parameterization-of-previous-studies}
    \end{small}
    \end{center}
    \vspace{-3ex}
\end{table*}
\renewcommand{\arraystretch}{1.0}

\vspace{1ex}

\noindent \textbf{\textit{R} factor: auxiliary resources}
The \textit{R} factor is the set of resources other than $\mathcal{M}$ that are available to the attacker and that can indirectly aid the disclosure of the target information. Possible examples are language models other than $\mathcal{M}$, corpora, and personal information of \textit{non-target} people.

Suppose again that the attacker guesses the diseases of the person $p_{1}$. An extreme example is that the model provider has released the model $\mathcal{M}$ whose pre-training data is not anonymized at all $(a=\mathrm{id})$, and the attacker has access to $\mathcal{D}_{\text{public}}$, the anonymized version of the pre-training data:

\vspace{-0.4ex}

\begin{center}
$
R\!=\!\{
\mathcal{D}_{\text{public}}\},\,
\mathcal{D}_{\text{public}}\!=\! a'(\mathcal{D}_{\text{private}}),\,
a'\!=\!f_{\text{HIPAA}}
$.
\end{center}

\vspace{-0.4ex}

\noindent In this scenario, $\mathcal{D}_{\text{public}}$ does not cause privacy leakage alone but may raise its risk because
the attacker only has to use $\mathcal{M}$ to fill $\mathcal{D}_{\text{public}}$ with the personal information removed from $\mathcal{D}_{\text{private}}$.

Another possible scenario is that the attacker has part of the pre-training data $\mathcal{D}_{\text{train}}^{\overline{\mathcal{P}}}$ where only non-target people are mentioned but cannot access the other part $\mathcal{D}_{\text{train}}^{\mathcal{P}}$:

\vspace{-0.5ex}

\begin{center}
$
R=\{
\mathcal{D}_{\text{train}}^{\overline{\mathcal{P}}}\}
$.
\end{center}

\vspace{-0.5ex}

\noindent In this scenario, the attacker may train a new language model $\mathcal{M}_{\text{shadow}}$ with $\mathcal{D}_{\text{train}}^{\overline{\mathcal{P}}}$ and attack $\mathcal{M}_{\text{shadow}}$ repeatedly to see how $\mathcal{M}_{\text{shadow}}$ reacts under the existence or absence of specific personal information in $\mathcal{D}_{\text{train}}^{\overline{\mathcal{P}}}$. This may enable the training of a classifier that can receive the reactions of $\mathcal{M}$ to the attacks and predict the existence or absence of specific target information in $\mathcal{D}_{\text{train}}^{\mathcal{P}}$.

Even without $\mathcal{D}_{\text{train}}^{\overline{\mathcal{P}}}$, similar attacks are possible if the attacker has a corpus $\widetilde{\mathcal{D}}_{\text{train}}$ that is irrelevant to $\mathcal{D}_{\text{train}}$ but has a very similar distribution:

\vspace{-0.5ex}

\begin{center}
$
R=\{\widetilde{\mathcal{D}}_{\text{train}}\}
$.
\end{center}

\vspace{-0.6ex}

Personal information can also be the \textit{R} factor. Suppose the attacker already knows the full name and diseases of non-target people $\overline{\mathcal{P
}}$, who are mentioned in $\mathcal{D}_{\text{private}}$ but not the target of the attack:

\vspace{-0.5ex}

\begin{center}
$
R=\{(\textit{full name}, \textit{diseases})_{\bar{p}}\}_{\bar{p}\in\overline{\mathcal{P}}}
$.
\end{center}

\vspace{-0.6ex}

This provides the attacker with positive samples of name-disease pairs in $\mathcal{D}_{\text{private}}$. The model $\mathcal{M}$ might act slightly differently to positive name-disease pairs and randomly generated negative name-disease pairs. Thus, the attacker may train a classifier to predict whether arbitrary name-disease pairs are present in $\mathcal{D}_{\text{private}}$~\cite{bert-privacy-naacl-2021-lehman}. We assume that the prior knowledge on non-target people fits the \textit{R} factor better than the \textit{K} factor since it provides an indirect clue for the privacy attack unlike the knowledge on target people.

\vspace{1ex}

\noindent \textbf{Summary} In any privacy leakage scenario, the attacker attempts to obtain the target information $I_{T}$ and attacks the pre-trained language model $\mathcal{M}$ to obtain the personal information in the pre-training data $I_{\mathcal{D}_{\text{private}}}$
using the prior knowledge $I_{K}$ and resources $R$:

\vspace{-3.5ex}

\begin{center}
$
\begin{aligned}
&\mathcal{M},I_{K}, R
\xrightarrow{\text{attack}}
\widehat{I_{T}}\\
&\widehat{I_{T}}\!\subseteq\!\{
    \widehat{
        \textit{category}_{\textit{person}}
    }
    |
    \textit{category}_{\textit{person}} \!\in\! m(a(I_{\mathcal{D}_{\text{private}}}))
\}.
\end{aligned}
$
\end{center}

\vspace{-0.5ex}

Various privacy leakage scenarios can be represented as a combination $(I_{K},a,R,I_{T})$, where each value corresponds to the \textit{K}, \textit{A}, \textit{R}, and \textit{T} factors.

\section{KART-based Review of Related Work}
\label{sec:kart-parameterization-of-previous-studies}

We overview privacy leakage scenarios in prior studies with our proposed KART parameterization, which is outlined in Table \ref{tb:kart-parameterization-of-previous-studies}.

\citet{membership_inference_gpt_misra} and~\citet{membership_inference_seq2seq} dealt with membership inference. They assessed the risk of a language model disclosing whether a specific document was in the training data $(I_{T}=\{\mathbbm{1}[d\in\mathcal{D}_{\text{train}}]\}_{d\in\mathcal{D}})$.~\citet{membership_inference_gpt_misra} discussed a GPT-1 model pre-trained with a public corpus and fine-tuned with a private corpus.~\citet{membership_inference_seq2seq} examined a Transformer model pre-trained with a private corpus. Both studies simulated the case that the attacker can access part of the pre-training data $(R=\{\mathcal{D}_{\text{train}}^{\overline{\mathcal{P}}}\})$ and trains a classifier to distinguish documents in and out of $\mathcal{D}_{\text{private}}$.

\citet{exposure_and_unintended_memorization}, ~\citet{extracting_training_data_from_lm}, and ~\citet{training-data-leakage-analysis-in-language-models-inan} simulated NLG to restore substrings of the pre-training data without prior knowledge or auxiliary resources $(I_{K}=\varnothing, R=\varnothing)$. 
~\citet{exposure_and_unintended_memorization} provided a case study of risk assessment the disclosure of for credit card numbers or social security numbers.~\citet{extracting_training_data_from_lm} and~\citet{training-data-leakage-analysis-in-language-models-inan} considered the disclosure of any substring in the training data to be a privacy breach. That is, the attacker places no limits on the target information $(I_{T}=I_{U})$.

\citet{bert-privacy-naacl-2021-lehman} and~\citet{bert-privacy-vakili} assumed that an attacker cross-refers to multiple personal information to reveal diseases of patients. The attacker cannot learn about a target person if the model only outputs a disease, but the attacker can associate the disease with the person if the model relates the disease to another attribute such as a full name. This problem formulation has a drawback in that the range of the target information is limited. However, it is advantageous because it covers scenarios where the model does not output an exact substring of the pre-training data but a similar one, or where the model outputs multiple pieces of personal information non-adjacently within a sentence. Such privacy leakage has not been examined in other studies.~\citet{bert-privacy-naacl-2021-lehman} examined a scenario where the attacker already knows the full name and sex of the target people $(I_{K}=\{(\textit{full name}, \textit{sex})_{p}\}_{p\in\mathcal{P}})$ and uses them as an NLG prompt to disclose diseases $(I_{T}=\{\textit{diseases}_{p}\}_{p\in\mathcal{P}})$. They also simulated the case that the attacker knows the name, sex, and diseases of non-target people in the pre-training data $(R=\{(\textit{full name}, \textit{sex}, \textit{diseases})_{\overline{p}}\}_{\overline{p}\in\overline{\mathcal{P}}})$ and trains a classifier to predict whether arbitrary people appeared in the pre-training data.~\citet{bert-privacy-vakili} simulated predictions of the full name and diseases together $(I_{T}=\{(\textit{full name}, \textit{diseases})_{p}\}_{p\in\mathcal{P}})$ using no prior knowledge $(I_{K}=\varnothing)$ or the sex of the target people $(I_{K}=\{\textit{sex}_{p}\}_{p\in\mathcal{P}})$, although the risk was not directly compared.

This KART-based review shows that previous studies have covered various scenarios and attacks. These are difficult to directly compare at present but may be integrated in a future meta-analysis.

\section{Experimental Demonstration}
\label{sec:experiment}
We demonstrate scenario-aware privacy risk assessment and compare the privacy risk among scenarios under the same definition of privacy leakage.

\begin{figure*}[tb]
    \begin{center}
        \includegraphics[scale=0.30]{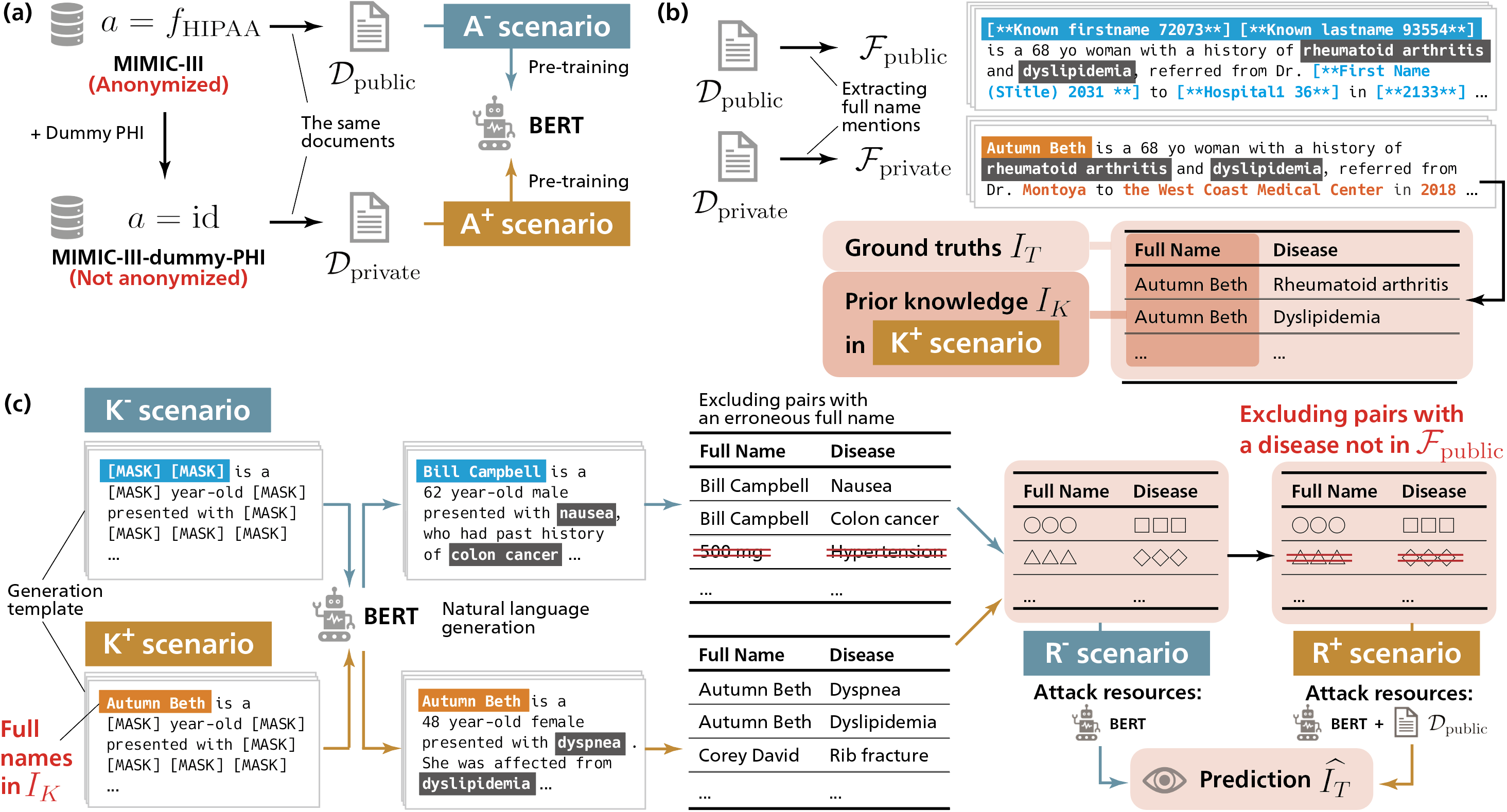}
        
        \vspace{-1ex}
    
        \caption{
        Overview of the privacy leakage experiment. (a) The model provider publishes a BERT model. Its pre-training data is anonymized in the $A^{-}$ scenarios $(\mathcal{D}_{\mathrm{public}})$ but not in the $A^{+}$ scenarios $(\mathcal{D}_{\mathrm{private}})$. (b) The attacker aims to reveal name-disease pairs present in ``full name mentions'' in $\mathcal{D}_{\mathrm{private}}$. (c) Attack with NLG using the pre-trained BERT model. Different templates are used in the $K^{+}$ and $K^{-}$ scenarios. Predictions are refined differently when $\mathcal{D}_{\mathrm{public}}$ is available ($R^{+}$ scenarios) or unavailable ($R^{-}$ scenarios) to the attacker.}
        \label{fig:privacy_attack}
    \end{center}
    \vspace{-2.5ex}
\end{figure*}

\subsection{Scenarios}
\label{sec:scenarios}
We prepared eight privacy leakage scenarios. In all the scenarios, a pre-trained BERT model $\mathcal{M}$ is published whose pre-training data comprises clinical records in a hospital. An attacker exploits $\mathcal{M}$ to estimate the full name and diseases of the patients in the pre-training data. The scenarios, however, have different details that may affect privacy risks. We name each scenario using ``\textit{K$^{+}$},'' ``\textit{K$^{-}$},'' ``\textit{A$^{+}$},'' ``\textit{A$^{-}$},'' ``\textit{R$^{+}$},'' and ``\textit{R$^{-}$}'' because they are represented by different \textit{K}, \textit{A}, and \textit{R} factors:

\vspace{-0.5ex}

\begin{center}
$
\begin{aligned}
    &\textit{K}^{+}: I_{K}= \{\textit{full name}_{p}\}_{p\in\mathcal{P}},
    &&\textit{K}^{-}: I_{K}=\varnothing, \\
    &\textit{A}^{+}: a= f_{\text{HIPAA}},
    &&\textit{A}^{-}: a=\mathrm{id},
    \\
    &\textit{R}^{+}: R=\{f_{\text{HIPAA}}(\mathcal{D}_{\text{private}})\},
    &&\textit{R}^{-}:  R=\varnothing
\end{aligned}
$
\end{center}

\vspace{-0.5ex}

\textit{X$^{+}$} and \textit{X$^{-}$} are higher and lower severity choices for the \textit{X} factor, respectively.

\vspace{1ex}

\noindent \textbf{\textit{K$^{+}$A$^{+}$R$^{+}$} scenario} Let $\mathcal{D}_{\text{private}}$ be the clinical records. They are used with no anonymization to pre-train a BERT model $\mathcal{M}$ from scratch $(\mathcal{D}_{\text{train}}=a(\mathcal{D}_{\text{private}}),\: a=\mathrm{id})$. In addition to $\mathcal{M}$, the model provider also publishes a corpus $\mathcal{D}_{\text{public}}$, which is composed of the same documents as $\mathcal{D}_{\text{private}}$ but anonymized under the HIPAA Privacy Rule $(\mathcal{D}_{\text{public}}=a'(\mathcal{D}_{\text{private}}),\: a'=f_{\text{HIPAA}})$. An attacker estimates the past or present diseases of $N$ patients $\mathcal{P}=\{p_{1},...,p_{N}\}$, all of whose full name is already known to the attacker. This scenario can be parameterized as follows: $I_{K}\!=\!\{\textit{full name}_{p} \}_{p\in\mathcal{P}},\:\:
a=\mathrm{id},\:\:R=\{\mathcal{D}_{\text{public}}\},$ and $
I_{T}=\{\textit{diseases}_{p}\}_{p\in\mathcal{P}}
$.

\vspace{1.2ex}

\noindent \textbf{\textit{K$^{+}$A$^{+}$R$^{-}$} scenario} The same as \textit{K$^{+}$A$^{+}$R$^{+}$} except that $\mathcal{D}_{\text{public}}$ is unavailable: $I_{K}\!=\!\{\textit{full name}_{p} \}_{p\in\mathcal{P}},\\
a=\mathrm{id},\:\:
R=\varnothing,$ and $
I_{T}=\{\textit{diseases}_{p}\}_{p\in\mathcal{P}}
$.

\vspace{1.2ex}

\noindent \textbf{\textit{K$^{+}$A$^{-}$R$^{+}$} scenario} The same as \textit{K$^{+}$A$^{+}$R$^{+}$} except that the pre-training data is anonymized under the HIPAA Privacy Rule: $I_{K}=\{\textit{full name}_{p} \}_{p\in\mathcal{P}},\:\:
a = f_{\text{HIPAA}},\:\:
R = \{\mathcal{D}_{\text{public}}\},$ and $
I_{T}=\{\textit{diseases}_{p}\}_{p\in\mathcal{P}}
$.

\vspace{1.2ex}

\noindent \textbf{\textit{K$^{-}$A$^{+}$R$^{+}$} scenario} The same as \textit{K$^{+}$A$^{+}$R$^{+}$} except that the attacker does not know the full name of the patients. Note that the attacker must guess the full name and diseases together since the prediction of the diseases alone does not often reveal the subject: $I_{K}=\varnothing,\:\:
a = \mathrm{id},\:\:
R = \{\mathcal{D}_{\text{public}}\},$ and $
I_{T}=\{(\textit{full name}, \textit{diseases})_{p}\}_{p\in\mathcal{P}}$.

\vspace{1.2ex}

\noindent \textbf{\textit{K$^{+}$A$^{-}$R$^{-}$} scenario} The same as \textit{K$^{+}$A$^{-}$R$^{+}$} except that $\mathcal{D}_{\text{public}}$ is unavailable: $I_{K}\!=\!\{\textit{full name}_{p} \}_{p\in\mathcal{P}},\\
a = f_{\text{HIPAA}},\:\:
R = \varnothing,$ and $
I_{T}=\{\textit{diseases}_{p}\}_{p\in\mathcal{P}}$.

\vspace{1.2ex}

\noindent \textbf{\textit{K$^{-}$A$^{+}$R$^{-}$} scenario} The same as \textit{K$^{-}$A$^{+}$R$^{+}$} except that $\mathcal{D}_{\text{public}}$ is unavailable: $
I_{K}=\varnothing,\:\:
a = \mathrm{id},\:\:
R = \varnothing,$ and $
I_{T}=\{(\textit{full name}, \textit{diseases})_{p}\}_{p\in\mathcal{P}}
$.

\vspace{1.2ex}

\noindent \textbf{\textit{K$^{-}$A$^{-}$R$^{+}$} scenario} The same as \textit{K$^{-}$A$^{+}$R$^{+}$} except that the pre-training data is anonymized under the HIPAA Privacy Rule: $
I_{K}=\varnothing,\:\:
a = f_{\text{HIPAA}},\:\:
R = \{\mathcal{D}_{\text{public}}\},$ and $
I_{T}=\{(\textit{full name}, \textit{diseases})_{p}\}_{p\in\mathcal{P}}
$.

\vspace{1.2ex}

\renewcommand{\arraystretch}{0.85}
\begin{table*}[tb!]
    \begin{center}
    \begin{small}
        \centering
        \caption{
        Results of the comparison of the privacy risk from a pre-trained language model in different scenarios.}
    \vspace{-2ex}
    \begin{tabular}{crrcrrr}\toprule
    \multicolumn{3}{c}{Anchor scenario}&
    \multicolumn{3}{c}{Weakened scenario}&
    \multirow{2}{*}{
            \makecell[l]{
                Privacy risk margin
            }
    }
    \\ \cmidrule{1-6}
        Name
    &
        \makecell[l]{
            Name-disease pairs\\
            per 10k generations\\
            (correct/valid)
        }
    &
        \makecell[l]{
            Privacy\\
            leakage\\
            ratio$^{*}$
        }
    &
        Name
    &
        \makecell[l]{
            Name-disease pairs\\
            per 10k generations\\
            (correct/valid)
        }
    &
        \makecell[l]{
            Privacy\\
            leakage\\
            ratio$^{*}$
        }
    &
    \\ \midrule
    \multirow{3}{*}{\textit{K$^{+}$A$^{+}$R$^{+}$}} &
    \multirow{3}{*}{351 / 10,049} &
    \multirow{3}{*}{3.50\%} &
    \textit{K$^{+}$A$^{+}$R$^{-}$} & 351 / 27,239 & 1.29\% &
    \textbf{2.21\%} \textbf{($\ge$0\%)}
    \\ 
    & & &
    \textit{K$^{+}$A$^{-}$R$^{+}$} & 306 / 9,721 & 3.14\% &
    \textbf{0.36\%} \textbf{($\ge$0\%)}
    \\ 
    & & &
    \textit{K$^{-}$A$^{+}$R$^{+}$} & 0 / 101 & 0.00\% &
    \textbf{3.50\%} \textbf{($\ge$0\%)}

    \\ \midrule
    
    \multirow{2}{*}{\textit{K$^{+}$A$^{+}$R$^{-}$}} &
    \multirow{2}{*}{351 / 27,239} &
    \multirow{2}{*}{1.29\%} &
    \textit{K$^{+}$A$^{-}$R$^{-}$} & 306 / 25,644 & 1.19\% &
    \textbf{0.10\%} \textbf{($\ge$0\%)}
    \\ 
    & & &
    \textit{K$^{-}$A$^{+}$R$^{-}$} & 0 / 249 & 0.00\% &
    \textbf{1.29\%} \textbf{($\ge$0\%)}
    
    \\ \midrule
    
    \multirow{2}{*}{\textit{K$^{+}$A$^{-}$R$^{+}$}} &
    \multirow{2}{*}{306 / 9,721} &
    \multirow{2}{*}{3.14\%} &
    \textit{K$^{+}$A$^{-}$R$^{-}$} & 306 / 25,644 & 1.19\% &
    \textbf{1.95\%} \textbf{($\ge$0\%)}
    \\ 
    & & &
    \textit{K$^{-}$A$^{-}$R$^{+}$} & 0 / 0 & NA$^{**}$ &
    NA$^{**}$

    \\ \midrule
    
    \multirow{2}{*}{\textit{K$^{-}$A$^{+}$R$^{+}$}} &
    \multirow{2}{*}{0 / 101} &
    \multirow{2}{*}{0.00\%} &
    \textit{K$^{-}$A$^{+}$R$^{-}$} & 0 / 249 & 0.00\% &
    \textbf{0.00\%} \textbf{($\ge$0\%)}
    \\ 
    & & &
    \textit{K$^{-}$A$^{-}$R$^{+}$} & 0 / 0 & NA$^{**}$ &
    NA$^{**}$

    \\ \midrule
    
    \textit{K$^{+}$A$^{-}$R$^{-}$} & 306 / 25,644 & 1.19\% &
    \textit{K$^{-}$A$^{-}$R$^{-}$} & 0 / 0 & NA$^{**}$ &
    NA$^{**}$

    \\ \midrule
    
    \textit{K$^{-}$A$^{+}$R$^{-}$} & 0 / 249 & 0.00\% &
    \textit{K$^{-}$A$^{-}$R$^{-}$} & 0 / 0 & NA$^{**}$ &
    NA$^{**}$

    \\ \midrule
    
    \textit{K$^{-}$A$^{-}$R$^{+}$} & 0 / 0 & NA$^{**}$ &
    \textit{K$^{-}$A$^{-}$R$^{-}$} & 0 / 0 & NA$^{**}$ &
    NA$^{**}$
    \\ \bottomrule
    \end{tabular}
    \label{tb:result}
    
    \vspace{-0.8ex}
    
    \begin{flushleft}
    $^{*}$Calculated with the actual numbers of pairs, not with the values in the table. $^{**}$No valid name-disease pairs were generated.
    \end{flushleft}
    
    \end{small}
    \end{center}
    
    \vspace{-2.5ex}
\end{table*}
\renewcommand{\arraystretch}{1.0}

\noindent \textbf{\textit{K$^{-}$A$^{-}$R$^{-}$} scenario} The same as \textit{K$^{-}$A$^{-}$R$^{+}$} except that $\mathcal{D}_{\text{public}}$ is unavailable: $
I_{K}=\varnothing,\:\:
a = f_{\text{HIPAA}},\\
R = \varnothing,$ and $
I_{T}=\{(\textit{full name}, \textit{diseases})_{p}\}_{p\in\mathcal{P}}
$.

\vspace{1ex}

This choice of scenarios is only an example and does not cover all real-world privacy leakage, but it may still help estimate the upper bound of risk since the \textit{K$^{+}$A$^{+}$R$^{+}$} scenario is favorable to the attacker.

For simplicity, we collectively refer to the scenarios with a parameter value $X^{+}$ or $X^{-}$ as ``$X^{+}$ scenarios'' or ``$X^{-}$ scenarios,'' respectively.

\subsection{BERT pre-training}
We pre-trained two BERT models, each of which was repeatedly used in $A^{+}$ and $A^{-}$ scenarios, respectively. Two sets of pre-training data were made by sampling the same 100k clinical records from  MIMIC-III~\cite{mimic-iii}, which comprises clinical records anonymized under the HIPAA Privacy Rule, and ``MIMIC-III-dummy-PHI,'' which we built by adding dummy personal information to MIMIC-III\footnote{The implementation is at \texttt{https://github.com/[*** masked ***]/[*** masked ***]}}. This method eliminated the risk of a real-world privacy breach. The samples from MIMIC-III-dummy-PHI corresponded to $\mathcal{D}_{\text{train}}$ in the $A^{+}$ scenarios and $\mathcal{D}_{\text{private}}$ in all the scenarios. The samples from MIMIC-III were used as $\mathcal{D}_{\text{train}}$ in the $A^{-}$ scenarios and $\mathcal{D}_{\text{public}}$ in the $R^{+}$ scenarios. See Appendices \ref{sec:mimic-iii-dummy-phi}, \ref{sec:appendix-bert-pretraining}, and \ref{sec:downstream-performance} for more details.

\subsection{Model inversion attack}

\subsubsection{Gold standard target information}
The privacy attack is a prediction of name-disease pairs $I_{T}\!=\!\{(n_{i}, d_{i,j})\}_{1 \le i \le N, 1 \le j \le N_{i}}$, where $n_{i}$ and $d_{i,1}, d_{i,2} , ...$ denote the full name and diseases of the $i$-th patient, respectively. We made the gold standard from $\mathcal{D}_{\text{private}}$ as in Figure \ref{fig:privacy_attack}. First, we extracted all ``full name mentions,'' five consecutive sentences beginning with the patient demographics ``(\textit{first name}) (\textit{last name}) is a (\textit{age}) year old (\textit{sex}).'' Then, we chose full name mentions $s_{1}, ..., s_{N}$ so that every full patient name in $\mathcal{F}_{\mathrm{private}}=\{s_{i}\}_{1 \le i \le N}$ was unique and encoded as two tokens by the BERT tokenizer. For each $s_{i}$, we extracted and normalized disease names into a controlled unique identifier (CUI) in the UMLS metathesaurus~\cite{umls} using MetaMap 2020~\cite{metamap}. We built $I_{T}$ by pairing a full name $n_{i}$ and CUIs $d_{i,1}, d_{i,2}, ...$ identified in $s_{i}$.

\begin{figure}[t]
    \begin{center}
        \includegraphics[scale=0.15]{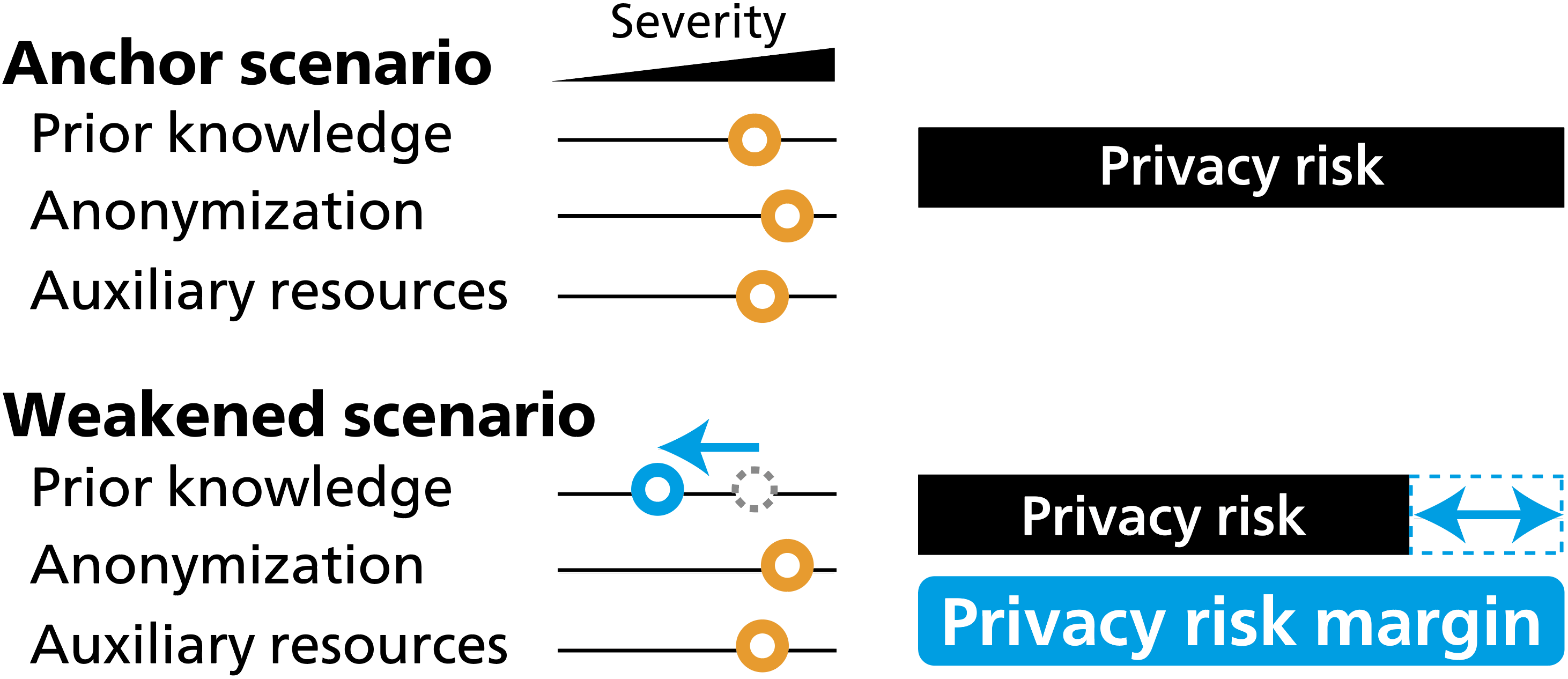}
        \vspace{-3.5ex}
        \caption{
        Risk comparison between anchor and weakened scenarios. All the primary factors of the weakened scenario are the same or less severe than those of the anchor scenario. The weakened scenario may result in a zero or positive privacy risk margin.
        }
        \label{fig:privacy_risk_comparison}
    \end{center}
    \vspace{-3.5ex}
\end{figure}

\subsubsection{Name-disease estimation}
To simulate attacks as in Figure \ref{fig:privacy_attack}, we generated at least 10k documents with the BERT model, extracted name-disease pairs, and made predictions $\widehat{I_{T}}$ by choosing ``valid'' pairs and excluding erroneous ones. Appendices \ref{sec:appendix-attack-detail} and \ref{sec:appendix-nlg-detail} give details.

\subsubsection{Risk comparison between scenarios}
We measured the privacy risk in each scenario with the \textit{privacy leakage ratio}, the ratio of the number of correct predictions to that of valid ones $(|I_{T} \cap \widehat{I_{T}}|\:/\:|\widehat{I_{T}}|)$. Then, we compared the risk in two scenarios differing in one factor as in Figure \ref{fig:privacy_risk_comparison}. For each pair, we referred to the more severe scenario as the \textit{anchor scenario} and the other as the \textit{weakened scenario}, and computed the \textit{privacy risk margin}, the pairwise difference of privacy leakage ratio.

\subsection{Results and analysis}
As shown in Table \ref{tb:result}, the privacy risk margin was greater than or equal to zero for all the scenario pairs where the margin could be calculated, suggesting that the scenario parameterizations coincided with the resulting privacy risk.

The upper bound of risk under this attack method could be approximated by the privacy risk ratio in the $\textit{K}^{+}\textit{A}^{+}\textit{R}^{+}$ scenario. Its magnitude may be small because the privacy leakage ratio in the $\textit{K}^{+}\textit{A}^{+}\textit{R}^{+}$ scenario may be mostly contributed to by random guesses rather than actual disclosures of personal information, given the small privacy risk margin between the $\textit{K}^{+}\textit{A}^{+}\textit{R}^{+}$ and $\textit{K}^{+}\textit{A}^{-}\textit{R}^{+}$ scenarios.

\section{Discussion}

We have introduced KART, a simple parameterization to clarify assumed privacy leakage scenarios during the risk assessment of sharing language models.

The estimation of the upper bound of privacy risk requires a wide coverage of real-world privacy leakage scenarios. Our KART-based review first simply clarified the scenarios dealt with in previous studies and their variety, suggesting the difficulty in direct comparison. Although we have not yet successfully integrated their findings, KART may provide a novel meta-analysis method for gaining comprehensive knowledge in the future. We have also shown that KART helps risk comparison in different scenarios under the same attack method or vice versa. Applying KART prior to every privacy leakage experiment would improve the comparability of future studies. KART should also spotlight scenarios that have not been fully explored.

The privacy risk margin was always zero or positive between anchor and weakened scenarios in our experiment. This may not always be consistent since privacy leakage occurs stochastically. However, if the scenario severity often coincides with the privacy risk margin in future studies, risk assessment may be streamlined by focusing on the most severe scenario possible under a given attack.

We assume that it is worthwhile simulating specific practical scenarios one by one in privacy risk assessment. KART alone does not provide a universal risk score valid in all scenarios. However, it has been unclear whether a single universal privacy risk metric covering all scenarios is possible. For example,~\citet{exposure_and_unintended_memorization} and~\citet{training-data-leakage-analysis-in-language-models-inan} proposed universal metrics of the upper bound of risk, but they focused on the disclosure of the exact substring of the pre-training data and did not cover other types of privacy leakage~\cite{bert-privacy-naacl-2021-lehman,bert-privacy-vakili}. Differential privacy~\cite{differential_privacy} is a strong framework to ensure privacy, but the relationship between $\epsilon$ and its real-world impact should still be examined~\cite{risk-evaluation-of-differential-privacy-emnlp-2021-hoory-shlomo}. Moreover, differential privacy in NLP is usually defined as the indistinguishability of two models pre-trained with datasets differing in one document. It is unclear whether the safety is robust to any privacy leakage, such as that caused by the generation of paraphrases of substrings of the dataset.~\citet{technical-privacy-metrics-survey-acm-2018-wagner-isabel} pointed out the existence of numerous privacy metrics and provided a guide to making suitable choices depending on the problem setting, which is probably a feasible means of privacy risk assessment.

It may be sufficient to always enforce a complete manual anonymization of pre-training data before sharing models for safe management. However, privacy risk assessment is still valuable, because language models can be accidentally made public if stolen or mistakenly put into public storage, as has happened to electronic health records~\cite{stolen_phi}. Moreover, the risk assessment may provide knowledge that may decrease future anonymization costs. Several off-the-shelf systems can effectively anonymize clinical records automatically~\cite{clinideid_is_the_best}, whose performance may be sufficient to publish language models pre-trained on anonymized corpora.

We expect KART to promote a wide range of future studies on privacy risks since it does not rely on domain-specific concepts.

\section{Conclusion}
It has been a challenge to assess the upper bound of the risk of sharing language models considering various real-world privacy leakage scenarios. We have proposed KART to simply parameterize complex scenarios. KART is expected to improve the portability of past and future privacy risk assessments and contribute to formulating privacy guidelines on language models.

\begingroup
{
\bibliography{reference}
\bibliographystyle{acl_natbib}
}
\endgroup

\clearpage

\appendix
\section{18 HIPAA Identifiers}
\label{sec:18-identifiers}

Under the HIPAA Privacy Rule, clinical records for the second usage must meet either of the two conditions: (i) Experts determine that the clinical records are anonymized properly and that there is little risk of disclosing the subject of the information, or (ii) a set of specific identifiers (18 HIPAA identifiers) regarding the subjects of the information and their relatives, employers, and household members is removed from the clinical records~\cite{hipaa}. Table \ref{tb:hipaa-18-identifiers} the lists 18 HIPAA identifiers.

\begin{table}[ht]
\begin{small}
\begin{center}
\caption{
18 identifiers to be masked under the HIPAA Privacy Rule.}

\vspace{-4ex}
\begin{tabular}[t]{|lp{6.35cm}|} \hline
    (A) &
    Names
    \\
    (B) &
    All geographic subdivisions smaller than a
    state, including street address, city, county,
    precinct, ZIP code, and their equivalent
    geocodes, except for the initial three digits of the ZIP code if, according to the current publicly available data from the Bureau of the Census: (1) The geographic unit formed by combining all ZIP codes with the same three initial digits contains more than 20,000 people and (2) the initial three digits of a ZIP code for all such geographic units containing 20,000 or fewer people are changed to 000
    \\
    (C) &
    All elements of dates (except year) for dates that are directly related to an individual, including birth date, admission date, discharge date, death date, and all ages over 89 and all elements of dates (including year) indicative of such age, except that such ages and elements may be aggregated into a single category of age 90 or older
    \\
    (D) &
    Telephone numbers 
    \\
    (E) &
    Fax numbers
    \\
    (F) &
    Email addresses 
    \\
    (G) &
    Social security numbers
    \\
    (H) &
    Medical record numbers
    \\
    (I) &
    Health plan beneficiary numbers
    \\
    (J) &
    Account numbers  
    \\
    (K) &
    Certificate/license numbers
    \\
    (L) &
    Vehicle identifiers and serial numbers, including license plate numbers
    \\
    (M) &
    Device identifiers and serial numbers
    \\
    (N) &
    Web Universal Resource Locators (URLs)
    \\
    (O) &
    Internet Protocol (IP) addresses
    \\
    (P) &
    Biometric identifiers, including finger and voice prints
    \\
    (Q) &
    Full-face photographs and any comparable images
    \\
    (R) &
    Any other unique identifying number, characteristic, or code, except as permitted
    \\ \hline
\end{tabular}
\label{tb:hipaa-18-identifiers}
\end{center}
\end{small}
\end{table}
\renewcommand{\arraystretch}{1.0}

\vspace{2ex}

\section{Details of Privacy Leakage Experiment}

\subsection{Pre-training Data}
\label{sec:mimic-iii-dummy-phi}
MIMIC-III~\cite{mimic-iii} is a publicly available dataset including over 2M clinical records of patients in the intensive care unit of Beth Israel Deaconess Medical Center. The clinical records are divided into 15 categories, including discharge summaries and progress notes. The clinical records are anonymized under the HIPAA Privacy Rule by replacing PHI incorporated into the HIPAA 18 identifiers with de-identification placeholders. To make MIMIC-III-dummy-PHI, we replaced the placeholders with dummy PHI. Dummy hospital names were randomly sampled from the i2b2 2006 dataset~\cite{i2b2_2006}, and the other dummy identifiers were randomly generated with \texttt{Faker}.\footnote{\texttt{https://github.com/joke2k/faker}}

Next, we built two sets of pre-training data by sampling the same documents from MIMIC-III-dummy-PHI and MIMIC-III. The sampling was a random choice of 50\% of the clinical records and further extraction of discharge summaries and progress notes, which left us with around 100k documents.

\subsection{BERT Pre-training}
\label{sec:appendix-bert-pretraining}
We pre-trained an uncased BERT-base model from scratch using $\mathcal{D}_{\text{train}}$. We did not fine-tune the BERT model provided by~\citet{bert}, which was pre-trained with BooksCorpus and Wikipedia, in order to avoid noise in the privacy leakage experiment. This was because it would be difficult to determine whether the full names and disease names output by the BERT model were disclosures from clinical records or just a reproduction of BooksCorpus or Wikipedia.

$\mathcal{D}_{\text{train}}$ was preprocessed in almost the same way as ClinicalBERT~\cite{clinicalbert}, but we did not delete digits. All the de-identification placeholders were removed.  Owing to limitations in computational resources, we pre-trained the BERT model for 1M steps with the maximum length set to 128. The other hyperparameters were the same as in ClinicalBERT: learning rate, 2e-5; batch size, 64.

\subsection{Privacy Attack Strategy}
\label{sec:appendix-attack-detail}

\noindent \textbf{Step 1: Name-disease pair generation} In the \textit{K$^{-}$} scenarios, the full patient name and diseases were estimated simultaneously. We generated $L$ documents $x_{1}, ..., x_{L}$ by filling \texttt{[MASK]} tokens of a 128-token-length template. The template contained four masking spans such as  ``\texttt{[CLS]} \texttt{(name-masking)} is a \texttt{(age-masking)} year old \texttt{(sex-masking)} presented with \texttt{(disease-masking)} \texttt{[SEP]}.'' In practice, the name-, age-, sex-, and disease-masking spans consisted of two, one, one, and 116 consecutive \texttt{[MASK]} tokens, respectively. Refer to Appendix \ref{sec:appendix-nlg-detail} for details of the method used to fill the blanks.

For each generated document $x_{l}$, we used the content filling the name-masking span as the prediction of the full patient name $\widehat{n_{l}}$. We also automatically extracted CUIs $\widehat{d_{l,1}}, \widehat{d_{l,2}}, ...$ using MetaMap 2020.  Finally, we collected name-disease pairs $\{(\widehat{n_{l}}, \widehat{d_{l,m}})\}_{1\le l \le L}$ from the $L$ generated documents.

In the \textit{K$^{+}$} scenarios, we obtained name-disease pairs similarly except that the name-masking span in the template was replaced with a randomly sampled full patient name $n_{i} \in I_{T}$ in each generation. This was because the attacker was supposed to already know the full patient names and only had to estimate the diseases.\\

\noindent \textbf{Step 2: Prediction refinement} The BERT output for the name-masking span sometimes made no sense as a person's full name. We excluded such ``invalid'' name-disease pairs where the predicted full name $\widehat{n_{l}}$ did not match any of the full names listed in the \texttt{Faker} library.

In the \textit{R$^{-}$} scenarios, the remaining ``valid'' name-disease pairs were used as the prediction $(\widehat{I_{T}})$:

\begin{center}

$
\begin{aligned}
\widehat{I_{T}} = \{(\widehat{n_{l}}
, \widehat{d_{l,m}}) \:|\: \widehat{n_{l}}
\in \textit{valid full names}
\}_{1 \le l \le L}
\end{aligned}
$.

\end{center}

In the \textit{R$^{+}$} scenarios, name-disease pairs were further excluded if their CUI had no corresponding disease name in $\mathcal{F}_{\mathrm{public}}$:

\begin{center}

$
\begin{aligned}
\widehat{I_{T}} = \{(\widehat{n_{l}}
, \widehat{d_{l,m}}) \:|\: \widehat{n_{l}}
\in \textit{valid full names}, \\ 
\widehat{d_{l,m}}\in\mathcal{F}_{\mathrm{public}}
\}_{1 \le l \le L}.
\end{aligned}
$

\end{center}

This is because the attacker, who has access to $\mathcal{F}_{\mathrm{public}}$, can assume that predictions are probably incorrect if they contain diseases that are absent from $\mathcal{F}_{\mathrm{public}}$.

Owing to limitations in computational resources and time, we obtained predictions for \textit{R$^{+}$} scenarios by refining corresponding ones for \textit{R$^{-}$} scenarios. For example, we made predictions $\widehat{I_{T}}$ for the \textit{K$^{+}$A$^{+}$R$^{+}$} scenario by reusing
name-disease pairs obtained in Step 1 in the \textit{K$^{+}$A$^{+}$R$^{-}$} scenario and then following Step 2.

\subsection{Details of Natural Language Generation}
\label{sec:appendix-nlg-detail}
Our NLG method is based on the Markov chain Monte Carlo method following~\citet{bert-has-a-mouth}. First, we designated the positions in the template where the $\texttt{[MASK]}$ tokens are initially placed as ``writable positions.'' Then, we repeated 1,000 iterations to randomly select one of the writable positions and to overwrite the word in that place with a new word. The new word was chosen randomly on the basis of the distribution given by masked language modeling in the first 250 iterations (burn-in period). Subsequently, we used a top-100 sampling strategy by setting the probability to zero for all of the words outside the top-100 posterior probabilities. The batch size was set to 32.

\section{Complementary Results}
\label{sec:complementary-results}

\subsection{Performance of BERT Models in Downstream Task}
\label{sec:downstream-performance}

We examined the performance of the pre-trained BERT models used in our experiment in the MedNLI task~\cite{mednli} to evaluate how well they were pre-trained. We fine-tuned the two BERT models used in \textit{A$^{+}$} and \textit{A$^{-}$} scenarios and the off-the-shelf model released by~\citet{bert}.

For each model, we calculated the validation accuracy for each learning rate $\in \{2e-5, 3e-5, 4e-5, 5e-5\}$ and each number of epochs $\in \{2,3,4\}$, and used the combination that maximized the validation accuracy for the test set. The batch size was set to 16.

Table \ref{tb:downstream-performance} shows the results. The performance of the off-the-shelf BERT model was comparable to that reported by~\citet{publicly-available-clinical-bert-embbeddings}. Our BERT model in \textit{A$^{-}$} scenarios outperformed the off-the-shelf model and our BERT model in \textit{A$^{+}$} scenarios achieved similar performance. Our BERT models benefited from being pre-trained with MIMIC-III, the source from which the premise sentences in MedNLI were extracted. However, our models are the same as the off-the-shelf model in that they are pre-trained for 1M steps from scratch and are disadvantageous for a much smaller pre-training corpus (120M vs 3,300M words). We assume that the BERT models used in our privacy leakage experiments are well pre-trained.

\begin{table}[ht]
    \begin{center}
    \begin{small}
        \centering
        \caption{
        Performance of the BERT models used in our experiment and the model provided by~\citet{bert} in the MedNLI test set.}
    \vspace{-1ex}
    \begin{tabular}{crrr}\toprule
    
    \multirow{2}{*}{Model}
    &
    \multicolumn{2}{c}{Hyperparameters}
    &
    \multirow{2}{*}{Test acc.}

    \\ \cmidrule{2-3}
    &
    Learning rate &
    Epoch &

    \\ \midrule
    Ours (\textit{A$^{+}$} scenarios) & 3e-5 & 4 & 72.29\%

    \\ 
    
    Ours (\textit{A$^{-}$} scenarios) & 2e-5 & 3 & 77.50\%

    \\ 
    
    \citet{bert} & 3e-5 & 3 & 76.09\%

    \\ \bottomrule
    \end{tabular}
    \label{tb:downstream-performance}
    
    \vspace{-0.8ex}
    
    \end{small}
    \end{center}
    
\end{table}

\renewcommand{\arraystretch}{0.95}
\begin{table}[ht]
    \begin{center}
    \begin{small}
        \centering
        \caption{
        Actual numbers of generated documents, valid name-disease pairs, and correct name-disease pairs in each scenario.}
    \vspace{-1ex}
    \begin{tabular}{crrr}\toprule
    
    \multirow{2}{*}{Scenario}
    &
    \multirow{2}{*}{
        Generated documents
    }
    &
    \multicolumn{2}{c}{Name-disease pairs}

    \\ \cmidrule{3-4}
    &
    &
    Valid &
    Correct

    \\ \midrule
    \textit{K$^{+}$A$^{+}$R$^{+}$} & 10,016 & 10,065 & 352

    \\ 
    
    \textit{K$^{+}$A$^{+}$R$^{-}$} & 10,016 & 27,283 & 352

    \\ 
    
    \textit{K$^{+}$A$^{-}$R$^{+}$} & 10,016 & 9,737 & 306

    \\ 
    
    \textit{K$^{+}$A$^{-}$R$^{-}$} & 10,016 & 25,685 & 306

    \\ 
    
    \textit{K$^{-}$A$^{+}$R$^{+}$} & 111,968 & 1,127 & 0
    
    \\ 

    \textit{K$^{-}$A$^{+}$R$^{-}$} & 111,968 & 2,789 & 0

    \\ 
    
    \textit{K$^{-}$A$^{-}$R$^{+}$} & 310,016 & 0 & 0
    
    \\ 
    
    \textit{K$^{-}$A$^{-}$R$^{-}$} & 310,016 & 0 & 0
    
    \\ \bottomrule
    \end{tabular}
    \label{tb:complementary_result}
    
    \vspace{-0.8ex}
    
    \end{small}
    \end{center}
    
\end{table}
\renewcommand{\arraystretch}{1.0}

\subsection{Privacy Leakage Experiment}

We show the actual numbers of generated documents, valid name-disease pairs, and correct name-disease pairs in each scenario in Table \ref{tb:complementary_result}, which is complementary to Table \ref{tb:result}. For the efficient use of computational resources, we increased the number of generations only for the four scenarios in which no correct name-disease pairs were obtained, but this did not change the result. See also Appendix \ref{sec:appendix-attack-detail}.

\end{document}